\documentclass[9pt]{IEEEtran}
\usepackage{graphicx}

\usepackage{fancyvrb}
\fvset{numbers=left,fontsize=\footnotesize}
\DefineShortVerb{\|}

\usepackage{paralist}

\begin{document}

\title{Modeling problems of identity in Little Red Riding Hood}
\author{
\IEEEauthorblockN{Ladislau B{\"o}l{\"o}ni}\\
\IEEEauthorblockA{
Dept. of Electrical Engineering and Computer Science\\
University of Central Florida\\
Orlando, FL 32816--2450\\
lboloni@eecs.ucf.edu
}
} 

\newcommand{\Xapagy}{Xapagy }
\newcommand{\Xapi}{Xapi }

\maketitle
\begin{abstract}

  This paper argues that the problem of identity is a critical challenge in agents which are able to reason about stories. The \Xapagy  architecture has been built from scratch to perform narrative reasoning and relies on a somewhat unusual approach to represent instances and identity. We illustrate the approach by a representation of the story of Little Red Riding Hood in the architecture, with a focus on the problem of identity raised by the narrative.

\end{abstract}


\section{Introduction}

\begin{figure}
\begin{center}
    \includegraphics[width=\columnwidth]{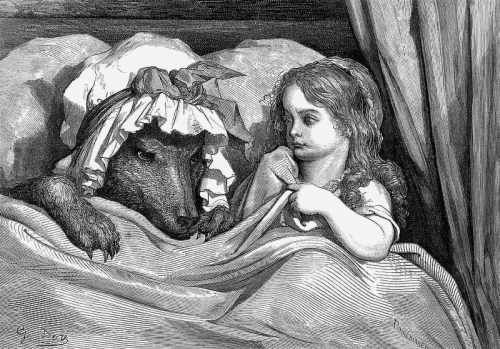} 
    \caption{\label{fig:LRRH} A case of confused identities: Little Red Riding Hood and the Big Bad Wolf (engraving by Gustave Dor{\'e}). 
}
\end{center}
\end{figure} 

{\bf Note:} To comply with the blind reviewing guidelines, the name of the system in this paper has been changed to SWNN (system with no name) and the name of the language employed by the system to LWNN (language with no name).

\medskip

We are swimming in a sea of stories, coming from printed, audio and visual media as well as delivered by live speech. Even more important is the narrative of our own lives, which includes events which we witness, but also stories we plan, infer, imagine or daydream.  

Agents interacting with humans will need to become adept on manipulating stories. This includes creating stories from their life experience, recalling or re-narrating stories with various levels of accuracy, predicting future events in stories, expressing surprise and so on. Some of this qualifies as {\em story understanding}, but we feel the term must be applied with care. Even for our own life narrative, there are many stories which we do not understand in all their deep implications, and many narratives we encounter are illogical, incoherent or lack deeper meaning. Human interactions proceed quite well without perfect  understanding. Some level of narrative reasoning ability, however, is necessary.

The \Xapagy cognitive architecture has been designed with the explicit goal to model and mimic the activities performed by humans when witnessing, reading, recalling, narrating and talking about stories. In contrast to other systems which treat the textual form of a narrative as a standalone, first class object, \Xapagy is an agent oriented system: a narrative is always seen through the experience of an individual agent, and the mode in which the narrative is delivered to the agent (for instance, the pace of story-telling or rhetorical breaks) changes the way in which the agent ``understands'' the story.

\Xapagy  has been developed from scratch, which required us to revisit many of the problems identified in the classic literature of the story understanding. Some of the representational decisions of the system, however, led to a number of {\em priority reversals}. In particular, the problem of identity became a determining factor of the overall architecture of the system. On the other hand, challenges such as anaphora resolution or the logical and/or psychological justification of actions had been relegated to those which we expect to be handled by the emergent properties of the system. 

This paper illustrates the problem of identity and the architectural solution for it offered by the \Xapagy  architecture. We shall use as a running example a relatively complex narrative, the story of Little Red Riding Hood (LRRH) and the Big Bad Wolf. This story is one of the archetypal fairy tales of the western culture. Almost everyone is exposed to it in early childhood. In many countries, it is the first middle length story a child is told. In contrast to the feel-good nature of modern fairy tales, the story is very dark, involving the gruesome (and in some versions, final) death of the three central characters. The story also has well known sexual connotations - in its original form being probably a cautionary tale for young girls against sexual predators. Overall, it exhibits a strong ``canonical strangeness'' in the definition used by Harold Bloom \cite{Bloom-1995-WesternCanon}.

LRRH poses a formidable series of problems from the point of view of computational narrative analysis. For systems which try to reason about stories based on the rational or psychologically justified actions of their participants, LRRH is a challenge, as the participants act in an irrational and psychologically unjustifiable way. There are internal logical gaps: we need to accept the fact that the wolf is able to eat Grandma and LRRH {\em because} LRRH talked to the wolf. There are physical impossibilities (the wolf swallowing Grandma and LRRH whole), as well as biological ones (they emerge unharmed from the wolf's belly).

Do children really {\em understand} LRRH? Parents often try to present the text as some kind of child-disobeying-parent educational message, without referring to the sexual connotations of the story. There is an example of Marxist literary criticism which presented the story as an allegory of class warfare.  

While creating a representation of the story which can be understood by a \Xapagy  agent, we have found that many (although not all) the difficulties are related to identities\footnote{We believe that using LRRH to illustrate the problems of identity makes the case better than stories in which the problem of identity is more blatant -- such as Dr. Jekyll and Mr. Hyde.}. For example: how does the LRRH in the mother's instructions relate to the real LRRH? Who does LRRH really talking to when asking Grandma about the size of her eyes? After we have imagined the gory death of Grandma and LRRH, are the persons emerging from the wolf's belly identical to the ones which have been swallowed? 

The remainder of this paper attempts to answer some of these questions by outlining key points of the representation of the LRRH story for the \Xapagy  agent. Section~\ref{sec:Informal} presents an informal introduction of the \Xapagy  system and the \Xapi pidgin language. In Section~\ref{sec:CaseStudies} we consider selected snippets from the fairy-tale, discuss the representational difficulties they represent and show how they can be converted to the \Xapi language. Related work is discussed in Section~\ref{sec:RelatedWork} and we conclude in Section~\ref{sec:Conclusions}.

\section{An informal introduction to \Xapagy }
\label{sec:Informal}

%
%

\subsection{External look: the pidgin language}

The \Xapagy  architecture describes the operation of an autonomous agent
which can directly witness events happening in the world, and it can communicate with humans and other agents through the {\em \Xapi pidgin language}. Pidgin languages \cite{sebba1997contact} are natural languages with a simplified syntactic structure which appear when two groups of people need to communicate without the time necessary to properly learn each other's languages. \Xapi shares some important features with human pidgin languages. It has an uncomplicated causal structure, it uses separate words to indicate degrees of properties, it has a fixed word order and does not support quantifiers. A line of \Xapi text represents a single sentence, with the sentence parts separated by ``/'' and terminated with a period ``.'' or question mark ``?''. 
\begin{quote}
\begin{Verbatim}
The girl / hits / the wolf.
Wh / eats / "LRRH"?
\end{Verbatim}
\end{quote}

The \Xapi parser translates a \Xapi statement into a single verb instance (VI) while compound statements into a several interconnected VIs. \Xapi supports a single form of compound sentence, the {\em quotation sentence}: 
\begin{quote}
\begin{Verbatim}
"LRRH" /says in "conversation" //
   eyes --of-- you / is-a / big.
\end{Verbatim}
\end{quote}

In some cases, the semantics of other compound or complex sentences can be approximated by sentences which refer to shared instances or VIs. We make, however, no claim that the expressive power of \Xapi matches that of a natural language.  

Subjects and objects are {\em instances} which are either currently in the {\em focus}, or are newly created by the sentence. A new instance can be created by prefixing a word with the indefinite article ``a/an'':

\begin{quote}
\begin{Verbatim}
"LRRH" / has / a hood.
\end{Verbatim}
\end{quote}

In this example we assume LRRH has been referred to before, but the hood has been just introduced in the story. Subsequent references to the already introduced instance of the hood are prefixed with the definite article ``the'' (which can be omitted for proper nouns).

In pidgin, we refer to instances through one or more of their {\em attributes}. When we mention the attribute |[basket]|, the reference will be made to the strongest instance in the scene which has the given attribute. In some cases, such as quotations, the resolution process is performed in the scene specified by the |in scene| part of the sentence, which might or might not be the same as the scene of the inquit.

The verb word in a \Xapi sentence actually maps to a mixture (overlay) of {\em verb concepts} in the internal representation of the \Xapagy  agent. The composition of this verb overlay determines the relationship between the sentences. For actions such as ``hits'' or ``bites'', the relationship between the sentences is one of a {\em weak temporal succession}. There are some sentences which do not represent actions in time, and thus they are not connected by succession relationships. Examples are verbs which set attributes to instances or establish relationships between instances:

\begin{quote}
\begin{Verbatim}
"LRRH" / is-a / small girl.
"Grandma" / loves / the girl.
\end{Verbatim}
\end{quote}

%
%

\subsection{From words to concepts and verbs}

We have seen that the \Xapi pidgin uses a simplified syntax, but otherwise regular English words. The {\em dictionary} of the agent maps nouns and adjectives to {\em overlays of concepts} while verbs and adverbs are mapped to {\em overlays of verb concepts}. We will discuss concept overlays, as the verb overlays are very similar. 

An overlay is the simultaneous activation of several concepts with specific levels of energy. For instance the dictionary of a \Xapagy  agent might associate the word ``girl'' with the following overlay:

\begin{Verbatim}[numbers=none]
[human=1.0, female=1.0, young=0.5 small=0.5]
\end{Verbatim}

The attributes of an instance are represented by an overlay which can be gradually extended through the side effects of the sentences. Thus, when reading |"LRRH" / is-a / girl| the instance identified with the attribute LRRH will acquire the attributes described in the overlay: human, female, young and so on. 

Concepts are {\em internal} structures of the \Xapagy  agent. To distinguish them from {\em words}, which are external entities, we will always show them in brackets, such as |[female]|.

Concepts can {\em overlap} on a pair-by-pair basis. For instance, there is a full overlap between man and human, meaning all men are human: {\em overlap\,}(|[man]|,|[human]|) = {\em area\,}(|[man]|). Thus, if we  inquire whether LRRH is human, we shall obtain a value of 1.0. There is, on the other hand, only a partial overlap between courageous and fearless: {\em overlap\,}(|[fearless]|, |[courageous]|) = 0.5$\cdot${\em area\,}(|[courageous]|). 

Words denoting proper nouns, such as |"LRRH"|, marked in pidgin by quotation marks, are treated slightly differently: when the agent first encounters a proper noun, it will create a new concept with a very small area, and an entry in the domain dictionary associating the proper noun with an overlay containing exclusively the new concept. Other than this, proper nouns are just like any other attributes. Having the same proper noun as an attribute does not immediately imply any form of identity. 

The dictionary which maps from a word to an overlay, the areas and overlap of the concepts are part of the {\em domain knowledge} of the agent. Different agents might have different domain knowledge - thus the meaning of the word might differ between agents. 

%
%

\subsection{Instances}

The definition of an instance in \Xapagy  is somewhat different from the way this term is used in other intelligent systems. Instead of representing an entity of the real world, it represents an {\em entity of the story, over a time span limited by the additivity of the attributes}. For a particular instance, its attributes, represented in a form of an overlay of concepts, are additive: once an instance acquired an attribute, the attribute remains attached to the instance forever. 

The advantage of this definition is that once we have identified an instance, there is no need for further qualification in order to identify its attributes (nor its actions). 

What might be counter-intuitive for the reader, however, is that things we colloquially call a single entity are represented in \Xapagy  by several instances. Let us, for instance, consider LRRH. There are several versions of the story, ranging from Charles Perrault's (with no happy ending) and the widely known Brothers Grim version, to the countless adaptations and parodies in modern media, including the 2011 movie starring Amanda Seyfried. 

In \Xapagy , these are all different instances, which share the attribute |["LRRH"]|. These instances, can be connected through various {\em relations of identity} (somatic, psychological, analogical and so on). The \Xapagy  system, however, moves a step beyond this. Not only LRRH from the Brothers Grim story and LRRH from the Hollywood movie are represented by different instances, but LRRH the live girl and LRRH the food item in the wolf's belly are also two different instances, as the change can not be represented as an addition of attributes. 

Whenever, at a later time, the \Xapagy  agent recalls LRRH (for instance, in a conversation) it first needs to establish which instance is under consideration. Once this instance has been unequivocally established to be, for instance, the live LRRH at the beginning of the story, all the attributes of the instance are also unambiguously established: we can say that she is young, happy, naive etc., attributes which would not make sense applied to the LRRH-as-food instance.   

%
%

\subsection{The focus}

The focus in the \Xapagy  system holds instances and VIs after their creation for a limited time interval during which they are {\em changeable}. After an instance or VI leaves the focus, it can never return - and thus, it remains unchanged.

Instances in the focus can acquire new attributes, participate as a subject or object in VIs and become part of relations. VIs in the focus can become part of succession or summarization relations, and they can be referred to by new VIs. 

A visual thinking oriented reader might think about the focus in the following way: the focus is a dynamically evolving graph. New nodes (instances and VIs) are added through various events. The same events might also create new edges among the nodes of the focus. When a node leaves the focus, it retains its attributes and edges, but it can not acquire new ones any more. So the focus can be seen as the actively growing boundary of a graph which represents the complete experience of the \Xapagy  agent. The graph will be only {\em locally connected}: it will {\em not} have long links, as only nodes which have been in the focus together can have links. 

The instances and VIs participate in the focus with a dynamically evolving weight. In absence of any events, the weights are gradually decreasing. Instances are refreshed when they participate in new VIs (events). Action events are ``pushed out'' from the instance by their successors. In addition to these, the weights are affected by a number of other dynamic factors.

%
%

\subsection{Shadows and headless shadows}

Instances and VIs leaving the focus will be {\em demoted to the memory} of the \Xapagy  agent with a certain level of {\em salience}. They will never enter the focus again. On the other hand, each instance and VI in the focus has a {\em shadow}, a weighted collection of instances and, respectively, VIs from the memory. 

The shadows are maintained through a combination of techniques whose goal is to make the shadows consistent between each other and match the ongoing story with the collections of stories in the shadows. 

Shadows are always matched to instances and VIs in the focus. The VIs in shadows, however, bring with themselves VIs to which they are connected through succession, context and summarization relations. These VIs can be clustered into weighted sets which are very similar to shadows, but they are {\em not} connected to current focus components. These sets are called {\em headless shadows} and they represent outlines of events which the agent either expect to happen in the future or assume that they had already happened but have not been witnessed (or they are missing from the narration). If an event matching the headless shadow happens, the two are combined to become a regular focus-component / shadow pair. 

Shadowing is the fundamental reasoning mechanism of the \Xapagy  architecture. All the higher level narrative reasoning methods rely on the maintenance of the shadows. For instance, the \Xapagy  agent can predict that the wolf will eat LRRH using a headless shadow, and can express surprise if this does not happen. While in this example the shadow is created after the events are inserted into the focus from an external source (for instance, by reading), the opposite is also possible. In the case of recall, narration, or confabulation, the agent creates instances and VIs in the focus based on pre-existing headless shadows. 

%
%

\subsection{Diffusion activities, spike activities and elements of prosody}

The state of a \Xapagy  agent is modified by two kind of {\em activities}: {\em spike activities} (SA) and {\em diffusion activities} (DA). 

SAs are instantaneous operations on overlays and weighted sets. Examples of activities modeled by SAs include inserting an instance in the focus, inserting a VI in the focus, and enacting the side effects of a VI. SAs are not parallel: the \Xapagy  agent executes a single SA at a time. 

DAs represent gradual changes in the overlays and weighted sets; the amount of change depends on the amount of time the diffusion was running. Multiple DAs run in parallel, reciprocally influencing each other. As a practical matter the \Xapagy  system implements DAs through discrete steps, with a temporal resolution an order of magnitude finer than the arrival rate of VIs. 

%
%

\subsection{Identity relations}

In \Xapagy the same instance represents an entity only as long as its attributes are additive (that is, the addition of new attributes can be perceived as {\em discovery}). Many situations which in colloquial terms appear to be occurrences of the same entity, in \Xapagy are represented with two or more instances connected by identity relations. Three types of identity relations are currently implemented:

\medskip

\noindent{|somatic-identity|: } connects instances which share the same physical body at different points in time. Two instances connected with somatic identity can not be simultaneously in focus. The somatic identity relation is created automatically as a side effect of the |changes| verb, which also removes the old instance from the focus. Somatic identity also connects the instances corresponding of the same character as seen over the course of time in multiple distinct episodes. By extension, somatic identity is also used to re-connect characters when a longer story is revisited (e.g. when reading a book over the course of several days). 

\medskip

\noindent{|fictional-identity|:} used for the relationship between an instance, and alter-ego's of this instance in fictional pasts or futures. Examples involve planning, daydreaming, or lying. Instances connected with fictional identity can be simultaneously present in the focus, in which case they will automatically be in each other's shadows. 

\medskip

\noindent{|view-identity|:} used for instances which are viewed differently in different scenes. They represent different interpretations of the scene, such as in cases of role playing. View identity is inherited when one of the instances is replaced by a somatically identical new instance as an effect of the |changes| verb.

\section{Case studies of identity problems in LRRH}
\label{sec:CaseStudies}

As an exercise in story representation, we have translated the Brothers Grim version of the story into the \Xapi language. Experimentally, we also created several alternative, postmodern versions. Naturally, this required the creation of an adequate vocabulary, as well as concept and verb concept databases. As the reasoning process of the \Xapagy  agent relies on previous experience, in order to obtain a level of understanding comparable to, let us say, a four year old girl, we would need to provide it with the background of stories posessed by the four year old girl. The development of such a background story database is a major knowledge engineering problem, and our ongoing work will be reported elsewhere. 

During the translation process, we have found that LRRH has been represented by 6-7 different instances, making the definition and handling of identity one of the critical problems of representation and understanding. 

In the following we shall select snippets of the story in natural language and illustrate how the problems of identity were handled during the translation into the \Xapi format. 

%
%

\subsection{The existence and identity of the narrative voice}

\begin{quote} \em
A little girl, Cindy, goes to bed and asks her Daddy to read her a story. 

``I am going to read you a story, written by the Brothers Grim. Once upon a time, there was a little girl, who had a red riding hood.'' 

``I have a red hood myself'', says Cindy.

``They called her Little Red Riding Hood''.

\end{quote}

As short as it is, this little snippet introduces three narrative planes. The first is the bedroom, where Cindy is listening to a story narrated by her dad. The second is the one in which the Brothers Grim  are writing the story (presumably, by sitting at their desk, somewhere in Germany in the XVIII-th century - this is however, not explicit in the story). The third narrative plane involves the fairytale inhabited by LRRH.

The \Xapagy  system handles narrative planes by the system of {\em scenes}. A scene is a collection of instances in the focus. VIs usually refer to instances in the {\em current scene}. Quote type VIs refer to the current scene in the inquit and a specified scene in the quote. 
Scenes are not necessarily physical environments. We, however, hypothesize that in the evolution of human cognition, narrative planes have originated from the representation of physical scenes, thus they will tend to have relations between instances isomorphic to the physical world. This is largely the same assumption as the one advocated by George Lakoff \cite{Lakoff-1993-MetaphorTheory}. The \Xapagy  system does not currently provide a feature to automatically infer scenes. To simplify the problem of scene reference, in this paper we will assign a proper noun to every scene. 

With these preliminaries, the beginning of the snipped is translated to \Xapi as follows.

\begin{quote}
\begin{Verbatim}
A scene "bedroom" / is-current-scene.
A girl "Cindy" / exists.
"Cindy" / is-inside / a bed.
A big man / exists.
The man / is-parent-of / "Cindy".
A scene "writing"/ exists.
A scene "fairytale"/ exists.
Man --parent-of-- "Cindy"/ says in "writing"// 
   "BrothersGrim"/ exists.
Man --parent-of-- "Cindy"/ says in "writing"// 
   "BrothersGrim"/ writes in "fairytale"//
   A little girl/ exists.
\end{Verbatim}
\end{quote}

The three-level indirection in the last sentence might come as a surprise. Yet, from the perspective of the reader of this paper, one would have even more levels of indirection:

\begin{quote}
\begin{Verbatim}
I / read in "AAMAS-proceedings" //
author / writes in "bedroom" //
man --parent-of-- "Cindy"/ says in "writing"// 
"BrothersGrim"/ write in "fairytale"//
a little girl/ exists.
\end{Verbatim}
\end{quote}

Will the entire story be narrated with this level of indirection? We could, of course, use a macro to simplify the \Xapi input. We believe, however, that human listeners remove the levels of indirect narration from a longer story they follow, and eventually reach a point where they follow the story either through a single indirection or through none, as if directly witnessed (possibly this latter corresponds to the phenomena commonly known as {\em suspension of disbelief}). This simplification of narrative planes is represented in the \Xapi text. 

The first to disappear is the Brothers Grim scene, and Father becomes the direct narrator of the fairytale\footnote{Having multiple layers of indirections is a well known literary device, from the Arabian Nights to  Chaucer and Boccaccio. Some modern novels, such as Umberto Eco's ``The Name of the Rose'', use a staggering number of narrative planes to frame the story. However, a reader continuing the reading the next day will obviously not recreate (or even remember) the number of indirections. One narrative plan (the old monk writing in the monastery of Melk) will return at the end, but not the other ones.}. 


\begin{quote}
\begin{Verbatim}
Man/ says in "fairytale" //
   The girl/ has / a red hood.
"Cindy"/ says in "bedroom" // 
   I/ have/ a red hood.
Man / says in "fairytale" // 
   She / is-a / "LRRH".    
\end{Verbatim}
\end{quote}

A typical human reader will infer from the text that Cindy identifies herself with the girl in the story. Analogously, a \Xapagy  agent will automatically place Cindy in the shadow of the LRRH instance (and vice versa) due to the perceived similarities. This shadowing allows us to infer things in the future - for instance the fact that the emotional states of Cindy might mirror those of LRRH. Shadows, however, do not become part of the remembered stories. It is likely (but not guaranteed) that the shadows will be recreated in a roughly similar way when recalling the story. If this subconscious identification is made explicit, it can be represented through a {\em fictional identity} relation. 

\begin{quote}
\begin{Verbatim}
"Cindy" / thinks in "bedroom" // 
  I / is-fictionally-identical / 
  "LRRH" --of-- "fairytale".
\end{Verbatim}
\end{quote}

We can insert this statement explicitly in the narration, or a \Xapagy  agent's missing relation reasoning might infer it automatically (based on the strong presence of the shadows). Such a relation {\em will} become part of the episodic memory and guide future recalls. 

%
%

\subsection{Orders from Mom: identity in a hypothetical future}

\begin{quote} \em
    Little Red Riding Hood's mother told her: ``You will take a basket and fill it with bread, cheese, and a bottle of wine. You will take it to Grandma's house in the forest. ''
\end{quote}

The orders from mother are obviously a new narrative, which takes place in a different scene. We are concerned about (1) the relationship between the two scenes and (2) the relationship between LRRH who is present in both scenes.

\begin{quote}
\begin{Verbatim}
Woman --parent-of-- "LRRH"/ 
   implies in "fairytale"// 
   a scene "orders"/ exists.
She/ implies in "fairytale" //
  the "orders"/ is-future-hypothetical /
  the "fairytale".
She/ implies in "orders"// 
   a little girl/ exists.
She/ implies in "orders"// 
   the girl/ is-fictionally-identical / 
   the girl in "fairytale".
\end{Verbatim}
\end{quote}

It should not be surprising by now that the \Xapagy  system represents this with distinct scenes and distinct instances, which are connected with specific relations (|future-hypothetical| for the scenes and |fictional-identity| for LRRH). 

Having an identity relation only affects the way in which shadowing happens: instances connected with identity relations do not necessarily share attributes - a little girl can imagine an alternative reality in which she would be a princess, or a steam engine. Identity type relations, however, reinforce shadow matching, which allows the prediction mechanism to predict future actions (which might be possibly enacted) and the surprise mechanism to express surprise if the real continuation diverges from the predicted one.

%
%

\subsection{Telling to the wolf / Lying to the wolf}

In the story, LRRH meets the wolf and tells him about her plans. For illustrative purposes, we will discuss a variant of the story, where LRRH is actually lying about her plans.

\begin{quote} \em
``Where are you going, little girl?'', asked the wolf.
``I am going to a picnic with my boyfriend, the hunter. I am carrying a basket of food and wine, and a concealed weapon.''.
\end{quote}

\begin{quote}
\begin{Verbatim}
Scene / implies in "conversation" // 
  a girl / exists.
"LRRH"/ is-fictionally-identical/
  the girl -- in -- "conversation".
Scene / implies in "conversation" // 
  a wolf / exists.
The wolf / is-fictionally-identical/
  the wolf -- in -- "conversation".
Wolf / says in "conversation" //
   girl / goes-to / wh?
"LRRH"/ says in "conversation"// 
   I/ has /a basket.
"LRRH" / says in "conversation" // 
   I/ has /a weapon.
"LRRH" / says in "conversation" // 
   a man hunter / exists.
"LRRH" / says in "conversation" // 
   the hunter / is-boyfriend-of / I.  
\end{Verbatim}
\end{quote}

Let us see what exactly is going on here, and whether the arrangement will yield a behavior which mimics the human behavior about stories. At a surface level, we have a new scene, the conversation scene |"conversation"|, in which we have a new instance of a girl, fictional-identity connected to LRRH. The instances of the hunter boyfriend and the weapon are present only in this scene.

The \Xapagy  agent listening to this is a customer of the complete narrative, which started with Cindy and her dad in the bedroom. The agent can simply record this narration, setting up the relations as it goes. If at this moment the listener agent has a moment of respiro, it can enter into a confabulation mood, and it can generate a continuation of the story, for instance, from the perspective of the wolf:

\begin{quote}
\begin{Verbatim}
Wolf / thinks in "wolf-plan" //
  I / attacks / the girl.
Wolf / thinks in "wolf-plan" //
  The girl / shoots / I.
Wolf / thinks in "wolf-plan" //
  I/ changes /dead.
\end{Verbatim}
\end{quote}

\Xapagy  being strongly reliant on episodic memory, such a recall can only happen if the agent has previous experience of stories involving shooting and killing. 

%
%

\subsection{Swallowing grandma}

\begin{quote} \em
  The wolf knocked on the door. Grandma opened the door and the wolf gobbled her up!
\end{quote}

The surface meaning of the snippet can be trivially translated into \Xapi:

\begin{quote}
\begin{Verbatim}
The wolf / knocks / the door.
"Grandma" / opens / the door.
The wolf/ gobbles-up / "Grandma".
\end{Verbatim}
\end{quote}

What the \Xapagy  agent will understand of it, depends very much on the previous experience of the agent (we are assuming the vocabulary and life experience of a 4 year old) and how much time the agent has to infer missing actions. 


\medskip
\noindent {\bf Case 1:} If the agent has never encountered the word ``gobbles-up'' before, it will create a dictionary entry, and a new verb concept - but nor further inference will be made. The representation of the agent will not have grandma disappear from the scene. 

\medskip

\noindent {\bf Case 2:} The agent knows the word ``gobbles-up'' and equivalates it with ``eats''. The side effect of this verb will remove the object from the current scene, but other inferences depend on the stories in the episodic memory of the agent. If the agent has no experience in swallowing live food (nor in stories involving the concrete act of dying), it will not make the inference that Grandma is dead. If such experience exists, the inferred statements will be:

\begin{quote}
\begin{Verbatim}
"Grandma" / changes / not-alive. 
"Grandma" / changes / chewed-food.
The chewed-food / leaves-scene.
The wolf / is-a / not-hungry.
\end{Verbatim}
\end{quote}

These statements actually represent two instance replacements for the initial Grandma instance, which will be connected with the relation |somatic-identity|. 

These inferences will only be made if the agent can internally insert four VI's into the focus. This requires a {\em respiro} - either the pace of narration must be sufficiently slow, or the narrator might make a longer pause to allow the agent to fill in the gaps in the narrative. If the agent is reading, rather than listening to an ongoing narrative it can create its own respiro. If the pace of a narrative is very fast, the agent will still create headless shadows (which are maintained by DAs) but will never get to instantiate them (which requires triggering an SA). 

\medskip

\noindent {\bf Case 3:} If the agent which is listening to a narrative had heard or read the story of LRRH before, the strongest shadows will map to the previous instances of the story (even if there are some differences between the versions). In this case the strongest headless shadows will predict the continuation of the story along the lines of previous versions. The inferences made by the agent will most likely not be from its life experience, but from the previous versions of the story\footnote{It is a well known aspect of the children's story telling the child which jumps in with the inferred continuation of the story - such as in the famous beginning of Pinocchio: 

  {\em Centuries ago there lived---
  
``A king!'' my little readers will say immediately.   

No, children, you are mistaken. Once upon a time there was a piece of wood.}}.

%
%
\subsection{Impersonating grandma}

For mysterious reasons, the wolf chooses to impersonate grandma to the arriving LRRH. The scene evolves through the famous conversation:

\begin{quote} \em
``Why are your eyes so big?'' asked the girl.

``To see you better,'' said the wolf.

``But why are your ears so big?'' asked the girl.

``To hear you better,'' answered the wolf.

``But why is your mouth so big?'' asked Little Red Riding Hood.

``To swallow you better!'' said the wolf, and jumped up and swallowed Little Red Riding Hood whole.
\end{quote}

This scene makes very little sense from any narrative coherence point of view. It is hard to believe that LRRH mistakes the wolf for Grandma, but it is even harder to believe that the wolf chooses to act as if it would be Grandma. 

The \Xapagy  approach to represent this, when seen from the third person omniscient narration perspective (which is also the one of the Brothers Grim, the narrating father's and Cindy's) is to represent two separate scenes. One of them is the physical scene, inhabited by LRRH and the wolf, while the other one is the scene of the conversation, inhabited by an alternate version of LRRH and Grandma. The instances are linked together using |view-identity| relations. 

\begin{quote}
\begin{Verbatim}
A scene "GrandmasHouse" / is-only-scene.
A wolf / exists.
A little girl "LRRH" / exists.
A scene "conversation" / exists.
Scene "conversation" / is-current-scene.
An old woman "Grandma" / exists.
"Grandma" / is-view-identical 
   / the wolf -- in -- "GrandmasHouse".
A little girl / exists.
The girl / is-view-identical 
   / "LRRH" -- in -- "GrandmasHouse".
The girl / has-as-grandparent / "Grandma".
Scene "GrandmasHouse" / is-current-scene.
The girl / asks in "conversation"//
  eyes -- of -- "Grandma"/ wh is-a / big?
The wolf / says in "conversation"//
  eyes -- of -- "Grandma"/ sees good / the girl.     
The girl / asks in "conversation"//
   mouth -- of -- "Grandma"/ wh is-a / big?
The wolf / says in "conversation"//
   "Grandma" / swallows good / the girl.
The wolf / swallows / "LRRH".  
\end{Verbatim}
\end{quote}

One of the interesting consequences of this representation is that both scenes will create headless shadows which will predict different continuations. The conversation scene will predict actions appropriate to the Grandma-granddaughter relation, while the physical scene will predict actions relevant to the wolf-little girl relation. The Brothers Grim create suspense by choosing repeatedly from the continuations predicted by the conversation scene. At the moment when the wolf swallows LRRH, both the physical scene {\em and} the conversation scene created headless shadows which predicted the action.

%
%
\subsection{LRRH and grandma emerge from the wolf}

\begin{quote} \em 
    The hunter shot the wolf, and then cut his belly. Suddenly, Little Red Riding Hood emerged from the belly, followed by Grandma!
\end{quote}

The wolf will go through an identity change from a live animal to a dead one. New instances for LRRH and Grandma will be created as they appear from the wolf-s belly. The critical part of this snippet is that the narrator needs to establish a somatic identity relation between the newly created instances and the initial instances corresponding to those objects. 

\begin{quote}
\begin{Verbatim}
The hunter / shots / the wolf.
The wolf / changes / dead.
The hunter / cuts / the belly --of-- wolf.
An "LRRH" / exits / the belly.
"LRRH" / is-somatically-identical / 
   "LRRH" --in-- "Grandmashouse".
An "Grandma" / exits / the belly.
"Grandma" / is-somatically-identical / 
   "Grandma" --in-- "Grandmashouse".
\end{Verbatim}
\end{quote}

\section{Related work}
\label{sec:RelatedWork}

The \Xapagy  system, as a body of software implementation is positioned as a {\em cognitive architecture}, a software system which tries to model a large subset of human cognition\cite{Newell-1994-Unified}. The specific objective of the \Xapagy  architecture is to achieve narrative reasoning, {\em i.e.} to mimic the activities performed by humans when witnessing, reading, recalling, narrating and talking about stories. There is an immense body of literature relevant to this enterprise. We shall only mention some of them which had been influential in our position with regards to the problem of identity in \Xapagy .

%
%

\noindent{\bf The problem of personal identity in philosophy.} The ``Personal Identity'' entry in Stanford Encyclopedia of Philosophy \cite{StanfordEncycPhil-PersonalIdentity} mentions a number of topics on which philosophers had focused ranging from the self-definition of a person to the minimal conditions necessary to accept somebody as a person. What is relevant to the work at hand however, are the answers given to the problem of {\em persistence:} is a particular person identical to a person in a particular moment of the past?

The {\em Psychological Approach} assumes that a psychological relation between the two entities is necessary and/or sufficient to consider the two persons identical. A simplistic version of this approach is the {\em Memory Criterion}, a person is identical with a person in the past if it can remember the person's past experiences. Although the Memory Criterion is sometimes attributed to Locke\cite{Locke-Identity}, he was aware of its drawbacks and in fact anticipated many of the later objections. The greatest problem with the Memory Criterion is that memory is not perfect: an adult can remember events from childhood, and an elderly person might remember events from adulthood, but not from childhood (the {\em gallant officer's paradox}). 

Grice \cite{Grice-1941-PersonalIdentity} proposed a solution based on the idea of ``total temporal state'', which is essentially all the events a person remembers or can potentially remember at some moment in time. Overlap between the total temporal can prove the identity of two persons. A different approach proposed by Shoemaker\cite{Shoemaker-1984-89ff} is based on causal dependence: two persons are considered identical if they are connected by a continuous causal chain of psychological connections. To overcome the problem of imperfect memory both approaches are relying in fact on the transitive closure of a sparse graph, where the edges represent either remembrances or more general psychological connections.

The {\em Somatic Approach} ties the identity of the person to the physical object, the human animal which embodies it. The continuity of the persons are based on brute physical continuity.

A third view, sometimes called the {\em Simple View} denies that these or other identifiers of identity are necessary and sufficient. In this view, psychological or physical continuity are just probabilistic signs of the personal identity. The identity is a self-containing definition, it can not be decomposed or proved with a set of simpler concepts. In fact, the proponents of this view point out that there is no need for continuity of the person for the survival of the human animal.

One of the important issues related to the persistency problem is the problem of {\em fission}. In human terms, this is usually stated with a thought experiment of the brain of a person being replicated and embedded in the other person. This lead to a famous debate between Parfit and Lewis \cite{Parfit-Lewis-debate}.

The problem of identity in cognitive systems, with a special attention to the problem of narratives and philosophical implications had been treated in a series of papers by Nissan\cite{Nissan-2003-IdentificationI}.

%
%

\noindent{\bf Classical story understanding systems:} Story understanding as a separate body of work inside artificial intelligence has a rich tradition, starting from work by Charniak\cite{Charniak-1972-ChildrenStoryComprehension,Charniak-1983-PassingMarkers}, 
Schank and Abelson\cite{Schank-1977-Scripts}, Norvig\cite{Norvig-1989-MarkerPassing}, Ram\cite{Ram-1994-AQUA}, Narayanan\cite{Narayanan-1997-Thesis}, Mueller\cite{Mueller-2004-Understanding}. 

The term ``classic'' frequently refers to the fact that these approaches, although frequently relying on methods of learning, are not statistical in nature, and perform what, in recent the recent dispute of Norvig vs. Chomsky \cite{Norvig-2011-Chomsky} has become known as focusing on the {\em deep whys}. 

It is fair to say that classical story understanding models have been shadowed in the last decade by the statistics based NLP - even prompting some arguments that the statistical models represent the final solution for text understanding - a position we do not share. 

\Xapagy  is clearly a system which fits in this body of work: it is performing operations over narratives (stories), it attacks some of the same issues (such as the expression of surprise, unfolding the structure of the story etc), and it does not use statistical learning. On the other hand, \Xapagy  does not start from human language but from the relatively controlled \Xapi pidgin. 



%
%
\noindent{\bf Semantic NLP:} an extensive and varied class of recent systems are considering the semantics of natural language processing. 

The way in which the semantics is defined in these systems covers a large range. The {\em explicit semantic analysis} (ESA) approach proposed by Gabrilovich and Markovich \cite{Gabrilovich-2009-Wikipedia}. This work defines the semantics of a text by its position in a multi-dimensional semantic space, where each dimension is defined by a basic concept. The basic concepts, and their interdependencies are collected from the Wikipedia entries of the individual concepts.

Another class of semantic NLP systems perform language generation based on the observation of events in the real world, a type of systems which are frequently called {\em grounded language learning}. We shall consider as example the system described by Chen, Kim and Mooney \cite{Chen-2010-Sportscaster}. 

A body of work recently reported by Chambers and Jurafsky \cite{Chambers-2008-NarrativeEventChains, Chambers-2009-NarrativeSchemasParticipants} centers on the learning of {\em narrative schemas}, which they define as coherent sequences or sets of events with participants who are defined as semantic roles, which can be fitted by appropriate persons. 

\section{Conclusions}
\label{sec:Conclusions}

Using the concrete example of the story of Little Red Riding Hood, this paper argued that identity is one of the crucial problems in reasoning about narratives. We have also shown that the \Xapagy  architecture, with its unusual definitions with regard to instances, verb instances, concepts and scenes allow us to represent stories which pose significant representational difficulties in other systems. 

Normally, the page limitations of this article limited us to discuss only snippets of the fairy tale. The full implementation of the fairy tale in the \Xapi language is available for download from the website
|removed for anonymous review|.



\bibliographystyle{abbrv}
\bibliography{Xapagy,LRRH,PersonalIdentity,ClassicStoryUnderstanding}  

\end{document}